\documentclass{Interspeech}

\interspeechcameraready

\title{Exploring the Effect of Segmentation and Vocabulary Size \\ on Speech Tokenization for Speech Language Models}

\author[affiliation={1}]{Shunsuke}{Kando}
\author[affiliation={1}]{Yusuke}{Miyao}
\author[affiliation={2,1}]{Shinnosuke}{Takamichi}

\affiliation{Graduate School of Information Science and Technology}{The University of Tokyo}{Japan}
\affiliation{Faculty of Science and Technology}{Keio University}{Japan}
\email{\{skando,yusuke\}@is.s.u-tokyo.ac.jp, shinnosuke\_takamichi@keio.jp}


\usepackage{comment}
\usepackage{cite}
\usepackage{url}
\usepackage{float}
\usepackage{booktabs}
\usepackage{makecell}
\usepackage{multirow}
\usepackage{framed}
\usepackage{amssymb}
\usepackage{pifont}
\usepackage{tikz}

\newcommand{\cmark}{\ding{51}}%
\newcommand{\xmark}{\ding{55}}%

\newcommand{\SIeq}[2]{\SI[parse-numbers=false]{#1}{#2}} 

\begin{document}

\maketitle

\begin{abstract}
The purpose of speech tokenization is to transform a speech signal into a sequence of discrete representations, serving as the foundation for speech language models (SLMs).
While speech tokenization has many options, their effect on the performance of SLMs remains unclear.
This paper investigates two key aspects of speech tokenization: the segmentation width and the cluster size of discrete units.
First, we segment speech signals into fixed/variable widths and pooled representations.
We then train K-means models in multiple cluster sizes.
Through the evaluation on zero-shot spoken language understanding benchmarks, we find the positive effect of moderately coarse segmentation and bigger cluster size.
Notably, among the best-performing models, the most efficient one achieves a 50\% reduction in training data and a 70\% decrease in training runtime.
Our analysis highlights the importance of combining multiple tokens to enhance fine-grained spoken language understanding.
\end{abstract}

\section{Introduction}

With the recent breakthroughs in large language models for textual natural language processing, speech language models (SLMs) have emerged as a new paradigm for spoken language processing~\cite{lakhotiaGenerativeSpokenLanguage2021,hassidTextuallyPretrainedSpeech2023,borsosAudioLMLanguageModeling2023,huWavLLMRobustAdaptive2024}.
SLMs are built by training language models on top of discrete speech representations (called ``discrete units'').
The process of converting a speech signal into a discrete unit sequence is called ``speech tokenization''.
Speech tokenization is typically performed by quantizing representations obtained from self-supervised learning (SSL) models~\cite{oordRepresentationLearningContrastive2019,baevskiWav2vec20Framework2020,hsuHuBERTSelfSupervisedSpeech2021}.
Leveraging the rich representations of SSL models, SLMs trained on these discrete units have demonstrated strong performance in zero-shot spoken language understanding (SLU)~\cite{nguyenZeroResourceSpeech2020}, spoken dialogue
~\cite{nguyenGenerativeSpokenDialogue2023}, speech-to-speech translation~\cite{leeDirectSpeechSpeechTranslation2022}, and other related tasks.

Various speech tokenization techniques have been proposed to enhance SLM performance.
Generative Spoken Language Modeling (GSLM) and its variants simply apply K-means clustering to the SSL model representations as is~\cite{lakhotiaGenerativeSpokenLanguage2021,kharitonovTextFree2022,hassidTextuallyPretrainedSpeech2023}.
However, since SSL model representations typically correspond to approximately \SI{20}{\milli\second} speech, the resulting discrete unit sequence tends to be long.
This severely affects the training of the Transformer-based language model~\cite{vaswaniAttention2017} as computation cost increases quadratically with respect to the sequence length.
Besides, previous study suggests that speech SSL representations primarily encode phonetic rather than semantic feature~\cite{choiSelfSupervisedSpeechRepresentations2024}, which might impair the capability of SLMs on a deeper understanding of spoken language.
To address these issues, previous research has invented speech tokenization techniques that segment input speech into fixed or variable-width units before discretization~\cite{algayresGenerativeSpokenLanguage2023,baadeSyllableLMLearningCoarse2024,choSylberSyllabicEmbedding2024}.
While segmentation reduces sequence length, it might cause a loss of information preserved in the original representation, and there is no clear agreement on the optimal tokenization scheme for this tradeoff and its reason.

\begin{figure}[t]
\centering
\includegraphics[width=.5\textwidth]{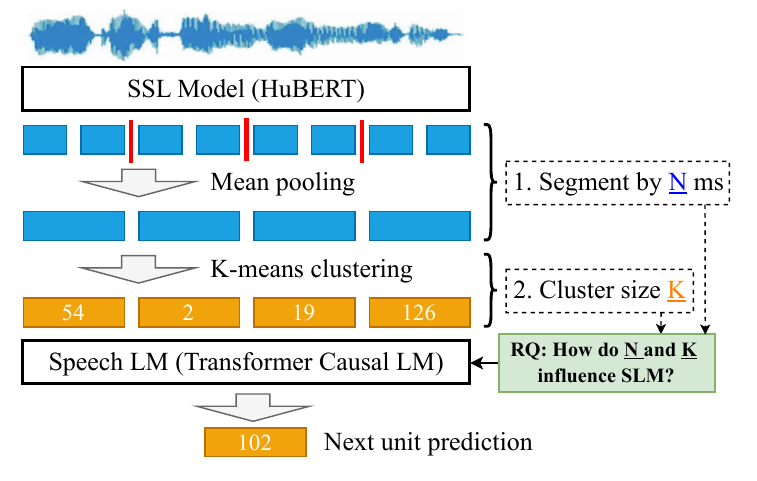}
\caption{
Overview of our research.
First, we extract continuous speech representation from the SSL model.
We add segments to the representation sequence by \SIeq{N}{\milli\second} and pooled them.
We apply K-means clustering to pooled representations with the cluster size of $K$.
By training SLMs in multiple settings of $N$ and $K$, we explore the optimal choice for spoken language understanding.
}
\label{fig:overview}
\end{figure}

This paper examines two key aspects of speech tokenization: the segmentation width and the cluster size of discrete units.
As depicted in Figure~\ref{fig:overview}, we first segment the SSL representation sequence in a fixed width and pooled features within each segment to obtain coarser representations.
Using these pooled representations, we then train K-means models to generate discrete unit sequences.
By applying multiple segmentation widths and varying the cluster sizes of the K-means model, we explore the optimal configurations for zero-shot SLU tasks.

Through comparative experiments, we find the positive effect of segmenting by moderately coarse width and making cluster size bigger at the same time.
We qualitatively suggest that larger segmentation width requires a larger vocabulary to accurately represent input speech.
Notably, a large segmentation setting reduces sequence length, enabling more lightweight training without sacrificing performance.
We also observe that specific benchmarks have different optimal settings, highlighting the importance of combining multiple tokens for SLU.
Besides fixed-width segmentation, we also investigate variable segmentation based on linguistic units (i.e., phonemes, syllables, and words) and compare their performances.
Our results demonstrate that variable segmentation does not show a clear advantage over fixed-width segmentation, suggesting that simpler segmentation methods may be preferable.

The experimental code is made publicly available\footnote{https://github.com/mynlp/speechlm}.

\section{Tokenization Methods to be Explored}
\label{sec:methods}

As depicted in Figure~\ref{fig:overview}, we first extract continuous speech representations of input speech using the SSL model.
Throughout this study, we used HuBERT~\cite{hsuHuBERTSelfSupervisedSpeech2021} as an SSL model and extracted representations from the ninth layer.
On top of this representation sequence, we performed speech tokenization in three steps.

\begin{enumerate}
\item Segment a sequence by \SIeq{N}{\milli\second} and apply mean pooling.
\item Apply K-means clustering with the cluster size $K$.
\item Deduplicate units. (e.g. 54 54 54 88 88 3 $\rightarrow$ 54 88 3)
\end{enumerate}

Since each HuBERT representation corresponds to \SI{20}{\milli\second} speech, $N$ is chosen as a multiple of 20.
We experiment with eight values: \{20, 40, 80, 120, 160, 200, 240, 280\}, where $N=20$ corresponds to the original sequence.
As $N$ becomes larger, the resulting sequence length becomes shorter by the factor of $N/20$.
As for $K$, the choice of value is not consistent across SLM studies, ranging from \{50, 100, 200\}~\cite{lakhotiaGenerativeSpokenLanguage2021} to \{5k, 10k, 20k\}~\cite{choSylberSyllabicEmbedding2024}.
To comprehensively cover this range, we experiment with eight values of powers of two: from $2^7=128$ to $2^{14}=16384$.
In total, we employed $8\times8=64$ tokenization methods.


Previous studies typically set a smaller cluster size for methods with small segment widths~\cite{lakhotiaGenerativeSpokenLanguage2021,borsosAudioLMLanguageModeling2023}, while larger segment widths are paired with larger cluster sizes~\cite{baadeSyllableLMLearningCoarse2024,choSylberSyllabicEmbedding2024}. 
This research supplements the cases of large/small cluster size and small/large segment width, aiming to gain deeper insight into these two aspects.

\section{Experimental Setup}

\subsection{Dataset}

\begin{table}[t]
\centering
\caption{
Minimum and maximum number of tokens for every segmentation width.
Minimum and maximum values correspond to $K=2^{7}$ and $K=2^{14}$, respectively.
}
\label{tab:dataset_stat}
\footnotesize
\begin{tabular}{l|p{0.54cm}p{0.42cm}p{0.42cm}p{0.42cm}p{0.42cm}p{0.42cm}p{0.42cm}p{0.42cm}}
\toprule
$N$ & 20 & 40 & 80 & 120 & 160 & 200 & 240 & 280 \\
\midrule
min & 87M & 63M & 39M & 27M & 21M & 17M & 14M & 12M \\
max & 127M & 77M & 42M & 28M & 22M & 17M & 14M & 12M \\
\bottomrule
\end{tabular}
\end{table}

As a training set for SLM, we used LibriSpeech~\cite{vassilLibrispeech2015}, a 960-hour English audiobook corpus.
Although this dataset is relatively small for SLM studies, our preliminary experiments showed that using a larger dataset (LibriLight~\cite{kahnLibriLight2020}; 60k hours audiobook corpus) did not lead to performance improvements.
A recent study on SLM based on syllable-level units~\cite{choSylberSyllabicEmbedding2024} also supports the use of LibriSpeech, as it reports better performance compared to baselines trained on the larger dataset.
Table~\ref{tab:dataset_stat} presents statistics on the training data.
As described in Section~\ref{sec:methods}, the sequence length is smaller when $N$ is larger, resulting in a smaller dataset size.
We show the minimum and maximum values across $K$.
The smaller $K$ is, the more likely there are repetitions, resulting in fewer tokens after deduplication.

\subsection{Model Setup}

We trained all K-means models on a 100-hour subset of the LibriSpeech training set.
For SLM training, we used OPT~\cite{zhangOPT2022}, a decoder-only Transformer language model.
We tuned hyperparameters to match GSLM~\cite{lakhotiaGenerativeSpokenLanguage2021}, resulting in 12 layers, 16 attention heads, embedding size of 1024, and FFN size of 4096.
To accelerate training, we concatenated all training data and grouped sequences into chunks of 2,048 tokens.
Each model was trained for up to 50,000 steps with a batch size of 16 on a single NVIDIA A100 GPU.
We applied an early stop when the validation loss did not improve for 1,000 consecutive steps.
We report the average scores of SLMs trained with three different random seeds.

\subsection{Evaluation}
\label{ssec:evaluation}

\begin{table}[t]
\centering
\footnotesize
\caption{Example pairs from benchmarks for spoken language understanding.}
\label{tab:benchmark}
\begin{tabular}{l|l}
\toprule
sBLIMP
&
\makecell[l]{
(\textcolor{green}{\textbf{\cmark}} Dogs eat meat,
\textcolor{red}{\textbf{\xmark}} Dogs eats meat)
}
\\
\midrule
sWUGGY
&
\makecell[l]{
(\textcolor{green}{\textbf{\cmark}} brick, \textcolor{red}{\textbf{\xmark}} blick)
}
\\
\midrule
pros-syntax
&
\makecell[l]{
\textcolor{green}{\textbf{\cmark}} But in the next breath [PAUSE] he cautioned. \\
\textcolor{red}{\textbf{\xmark}} \ But in the next [PAUSE] breath he cautioned.
}
\\
\midrule
pros-lexical
&
\makecell[l]{
\textcolor{green}{\textbf{\cmark}} But in the next [PAUSE] breath he cautioned. \\
\textcolor{red}{\textbf{\xmark}} \ But in the next breath he cau [PAUSE] tioned.
}
\\
\midrule
\makecell[l]{
tStoryCloze\\
(tSC)
}
&
\makecell[l]{
Ana was tanning on the beach. She dozed off in the \\ warm sun. She woke three hours later. Her eyes \\ widened as she looked in the mirror.\\ 
\textcolor{green}{\textbf{\cmark}} Ana was extremely sunburnt. \\
\textcolor{red}{\textbf{\xmark}} \ Michael hoped the new squirrel would fare.
}
\\
\bottomrule
\end{tabular}
\end{table}

We evaluate SLMs on five types of zero-shot SLU tasks shown in Table~\ref{tab:benchmark}.
Each task consists of pairs of correct and incorrect speech audio samples, and the model is evaluated based on its ability to assign a higher likelihood to the correct sample.
The chance rate is 0.5 for all tasks.

\textbf{sBLIMP}~\cite{nguyenZeroResourceSpeech2020} assesses the model’s grammatical knowledge.
Each task is categorized according to 12 types of linguistic phenomena, such as subject-verb agreement or argument structure.
\textbf{sWUGGY}~\cite{nguyenZeroResourceSpeech2020} verifies whether a model has lexical knowledge.
It consists of a pair of an existing word and a slightly modified nonce word. 
\textbf{pros-syntax} and \textbf{pros-lexical} are from prosaudit benchmark~\cite{deseysselProsAuditProsodicBenchmark2023}, which probes model's capability in handling prosodic information.
Stimulus pairs are constructed by inserting a \SI{400}{\milli\second} pause to the natural and unnatural position within speech.
In the pros-syntax task, the correct pause placement corresponds to a prosodic phrase boundary.
In the pros-lexical task, the correct placement is at a word boundary, while the incorrect placement is within a word.
\textbf{Topic SC (tSC)}~\cite{hassidTextuallyPretrainedSpeech2023} tests whether a model has commonsense knowledge.
This is a spoken version of StoryCloze~\cite{nasrinClozeEvaluation2016}, which rewrites the last sentence of a five-sentence story to produce an incoherent story.
Since the original Spoken SC (sSC) dataset is regarded as too challenging~\cite{hassidTextuallyPretrainedSpeech2023}, we used tSC instead, where the final sentence is randomly chosen from the dataset to generate topically incoherent story\footnote{
We also evaluated on sSC but found that all models performed at near-chance rate accuracy.
}.

\begin{figure*}[t]
\centering
\includegraphics[width=\textwidth]{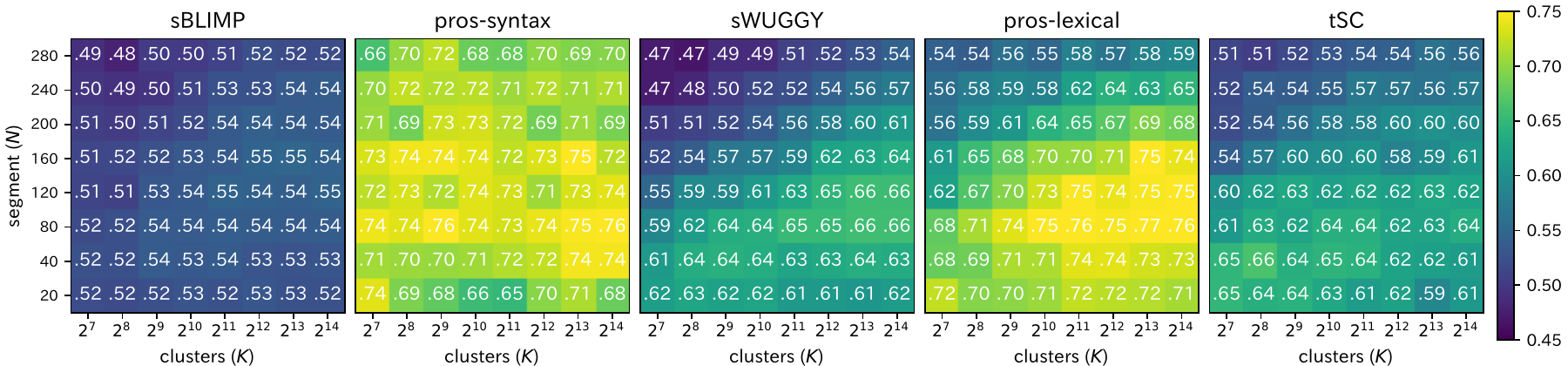}
\caption{
Main results of SLM performance on zero-shot SLU tasks.
}
\label{fig:main-result}
\end{figure*}

\begin{table}[t]
\centering
\caption{
Best performing $K$ values, average accuracies among five tasks, and training runtimes for $N=20, 40, 80, 120$.
}
\label{tab:train_stat}
\small
\begin{tabular}{l|lll}
\toprule
$N$ & Best $K$ & Avg. Acc. & Train Runtime \\
\midrule
20  & $2^7 \ \ \ (128)$  & 0.65 & 12.4 hours \\
40  & $2^{13} \ (8192)$  & 0.65 & 11.5 hours \\
80  & $2^{14} \ (16384)$ & 0.67 & 8.3 hours \\
120 & $2^{14} \ (16384)$ & 0.66 & 6.7 hours \\
\bottomrule
\end{tabular}
\end{table}

\section{Main Result}

For simplicity, we denote the configuration with segment width $N$ and cluster size $K$ as $(N,K)$.

Figure~\ref{fig:main-result} shows results on fixed boundary settings.
We observe that the best-performing configurations are centered around $(80, 2^{13})$.
An exception is tSC, where the optimal setting appears to be around $(40,2^8)$ (if any), though the differences in accuracy are not significant.
For a clear comparison, we identify the best $K$ values based on average accuracy across the five tasks.
We focus on relatively small $N$ values ($20,40,80,120$), as larger $N$ tends to degrade performance.
Table~\ref{tab:train_stat} shows the summary, including average training runtimes.
As we've seen in Table~\ref{tab:dataset_stat}, increasing $N$ results in smaller dataset size, which contributes to shorter training runtime.
Notably, the best-performing setting $(80, 2^{14})$ reduces the training data by 50\% (42M vs. 87M) and the training runtime by 70\% (8.3h vs. 12.4h) compared to $(20,2^7)$ setting.

In terms of benchmarks, while both sBLIMP and pros-syntax are related to syntactic knowledge, the accuracy on pros-syntax is significantly higher than on sBLIMP.
This suggests that SLMs have a high capability of handling prosodic features but struggle with a deeper understanding of natural language.
For pros-syntax, we observe exceptionally high accuracies even at the largest $N$ values.
This may be attributed to the fact that prosaudit inserts a \SI{400}{\milli\second} pause to stimuli, which is much longer than $N$.
For lexical tasks (sWUGGY and pros-lexical), although pros-lexical shows higher accuracy, both tasks exhibit similar overall trends.
On the other hand, tSC results show a slightly different tendency: there seems to be no clear optimal setting.
Investigating the underlying factors behind this difference remains for future research.

\section{Analysis}

\subsection{Effect of Larger $N$ and $K$}

\begin{figure}[t]
\centering
\includegraphics[width=.47\textwidth]{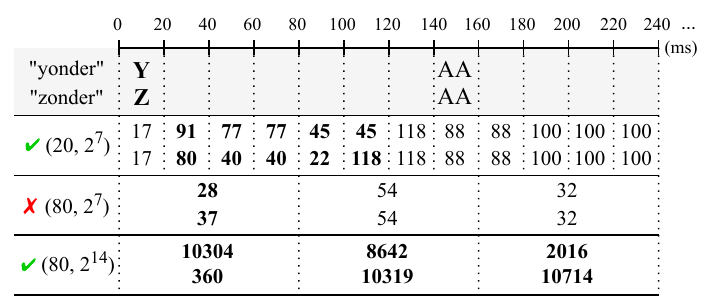}
\caption{
An example from sWUGGY where $(20, 2^7)$ and $(80, 2^{14})$ can solve but $(80, 2^7)$ fails.
Differences in phoneme or unit are shown in bold.
The first row shows the actual stimuli from the dataset and the rest shows unit sequences.
Since the dataset does not include phonetic alignments, we annotated them by ourselves using Praat~\cite{praat}.
}
\label{fig:error_analysis}
\end{figure}

Overall, for larger $N$, accuracy tends to improve with increasing $K$.
This observation is analogous to the relationship between phoneme and morphome: combining a small number of phonemes produces a large number of morphemes~\cite{martinet1966elements}.
In other words, in smaller $N$, the model does not require a large vocabulary because there are fewer categories that are essentially distinct.
As $N$ increases, the vocabulary size must also be larger to accommodate the growing number of categories.

To discuss it qualitatively, we extract cases where the combination of (large $N$, small $K$) fails but (small $N$, small $K$) and (large $N$, large $K$) can solve.
Figure~\ref{fig:error_analysis} shows an example from sWUGGY.
In this example, SLMs with setting $(20, 2^7)$ and $(80, 2^{14})$ could assign higher likelihood to the existing word ``yonder'', but $(80, 2^7)$ could not.
There is a difference in phoneme up to \SI{140}{\milli\second} (Y vs. Z), which is captured by settings $(20,2^7)$ and $(80,2^{14})$ as the discrete unit sequences differ within this range.
However, the setting $(80,2^7)$ fails to reflect the difference between \SI{80}{\milli\second} and \SI{140}{\milli\second}: in this range, it assigns the same unit ``54'' to both stimuli.
This might be attributed to the lack of vocabulary, which is resolved by increasing $K$ from $2^7$ to $2^{14}$.
It would be interesting to investigate whether this effect applies to larger $N$ with much larger $K$, but that could make training difficult for both the K-means model and SLMs.
Future work could investigate on training SLMs with continuous representations, which can be viewed as the limit of discrete representations~\cite{nguyenAreDiscreteUnits2022}.

\subsection{sBLIMP Accuracy Split by Task Type}

\begin{figure*}[t]
\centering
\includegraphics[width=\textwidth]{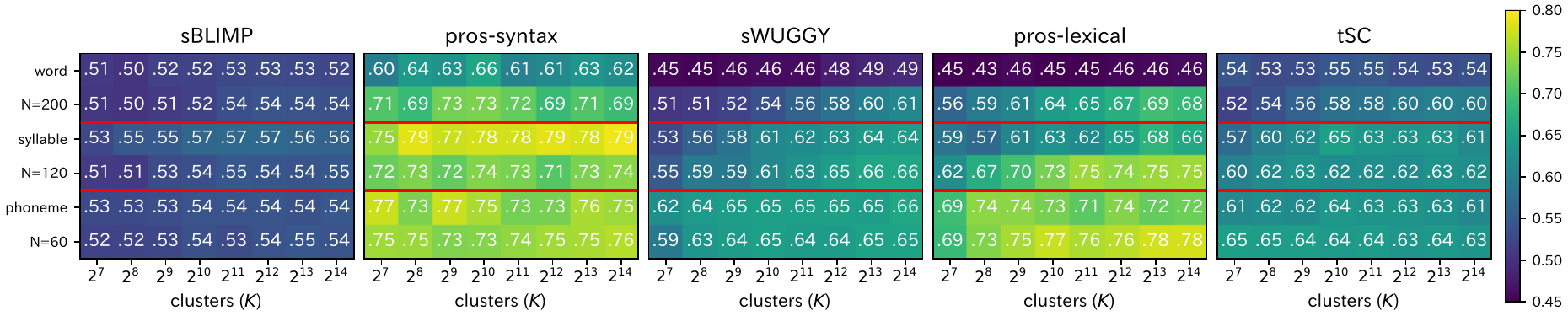}
\caption{
Results of variable segmentation on phoneme, syllable, and word levels.
For comparison, we show the fixed width segmentation results of which $N$ is the same as median of the distribution of variable segmentation.
}
\label{fig:pred}
\end{figure*}

\begin{figure}[t]
\centering
\footnotesize
\includegraphics[width=.47\textwidth]{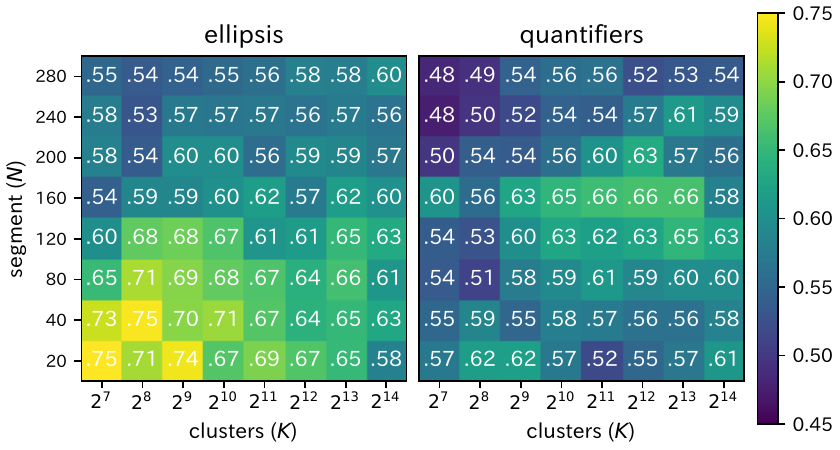}
\begin{tabular}{l|l}
\toprule
Task & Example \\
\midrule
ellipsis &
\makecell[{{p{5.5cm}}}]{
\textcolor{green}{\textbf{\cmark}} Anne’s doctor cleans one \underline{important} book and Stacey cleans a few. \\
\textcolor{red}{\textbf{\xmark}} Anne’s doctor cleans one book and Stacey cleans a few \underline{important}.
}\\
\midrule
quantifiers & 
\makecell[{{p{5.5cm}}}]{
\textcolor{green}{\textbf{\cmark}} No boy knew \underline{fewer than} six guys. \\
\textcolor{red}{\textbf{\xmark}} \ No boy knew \underline{at most} six guys.
}\\
\bottomrule
\end{tabular}
\caption{
sBLIMP accuracies split by task type.
We show two results (ellipsis and quantifiers) which display unique tendency.
}
\label{fig:split_blimp}
\end{figure}

Figure~\ref{fig:main-result} shows that sBLIMP accuracy is almost chance rate for all settings.
This is consistent with findings from previous studies: even much larger SLMs also struggle with sBLIMP~\cite{hassidTextuallyPretrainedSpeech2023,nguyenSpiRitLM2024}.
Still, since sBLIMP is a suite of 12 distinct tasks, some of them might be solvable to some extent.
We split the accuracy according to the task and found that it is actually the case.
We show two examples in Figure~\ref{fig:split_blimp}: ``ellipsis'' and ``quantifiers''.
``Ellipsis'' tests the possibility of omitting expressions from the sentence.
``Quantifiers'' assesses whether the quantifier is placed in the right position.
The results suggest that the best accuracy for both tasks is significantly above chance.
Notably, the optimal settings are unique for these tasks: they are located in the vicinity of $(40, 2^8)$ for ellipsis and $(160, 2^{12})$ for quantifiers.
This tendency is clearly different from other benchmarks shown in Figure~\ref{fig:main-result}.
This finding highlights the importance of combining different types of tokens to enhance SLU, which supports previous studies~\cite{borsosAudioLMLanguageModeling2023,jiatongMultiResolutionHuBERT2024}.

\subsection{Effect of Variable Segmentation Width}

While we have discussed the results of fixed-width segmentation, it is natural to segment speech into variable-width segments based on linguistic units, such as phonemes, syllables, and words.
Therefore, we trained SLMs on the variable segmentation predicted by unsupervised segmentation methods.
Although previous studies have partially attempted this approach~\cite{algayresGenerativeSpokenLanguage2023,baadeSyllableLMLearningCoarse2024,choSylberSyllabicEmbedding2024}, our goal is to investigate how different levels of linguistic units influence SLM performance under the comparative framework.
Also, since variable segmentation poses additional computation costs, we aim to assess whether it is beneficial by comparing it against the fixed-width setting.

We used UnsupSeg~\cite{kreukPhonemeSegmentation2020}, Sylber~\cite{choSylberSyllabicEmbedding2024}, and GradSeg~\cite{fuchsWordSegmentation2023} for segmenting speech into phoneme, syllable, and word units, respectively.
Similar to the fixed-width segmentation setting, we applied mean pooling to variable-width representations.
To compare against fixed-width segmentation settings, we computed medians of each segmentation width distribution\footnote{Since the distributions of segment width have a long tail, we used median instead of average as a representative value.}.
The medians for phoneme, syllable, and word segmentation were \SI{60}{\milli\second}, \SI{120}{\milli\second}, and \SI{200}{\milli\second}, respectively.

Figure~\ref{fig:pred} shows comparative results of fixed and variable segmentation settings\footnote{We additionally trained SLMs on $N=60$ setting for comparison.}.
We observe the positive effect of variable segmentation in limited settings: syllable segmentation on sBLIMP and pros-syntax.
Overall, the accuracies were comparable to those of fixed-width segmentation, even significantly impaired in the word segmentation settings.
As suggested in \cite{algayresGenerativeSpokenLanguage2023}, inaccurate segmentation may cause performance degradation.
Considering the computational cost of unsupervised segmentation, it might be more preferable to use fixed-width segmentation when training SLMs.
On the other hand, previous studies suggest learning syllable-level representations and using them instead of raw HuBERT representations for training SLMs, showing impressive performance in SLU tasks~\cite{baadeSyllableLMLearningCoarse2024,choSylberSyllabicEmbedding2024}.
Whether fixed or variable settings, future work could explore the benefits of learning segment-level representations for SLMs within our comparative experimental framework.

\section{Conclusion}

In this research, we explored the effect of speech tokenization on the SLU capabilities of SLMs.
We conducted multiple speech tokenizations based on the combination of the fixed/variable segmentation and the cluster size.
Our experiment on fixed-width segmentation suggests the positive effect of moderately coarse segmentation width and bigger cluster size, which contributes to a reduction in both training data size and runtime.
We find that the optimal tokenization settings vary across benchmarks, highlighting the importance of combining multiple tokens for further performance of SLMs.
We demonstrate that variable-width segmentations basically do not show a clear advantage over fixed-width segmentations.
While we conducted a comprehensive set of experiments on speech tokenization, the exact reasons why certain settings are optimal for each benchmark remain unclear.
Additionally, our focus was on SLU tasks, and we did not explore other areas such as speech synthesis or speech continuation.
We leave these explorations for future work.

\section{Acknowledgements}
This work was supported by
JST ACT-X JPMJAX24C9.

\bibliographystyle{IEEEtran}
\bibliography{mybib}

\end{document}